# Updating Sets of Probabilities

**Adam J. Grove**
NEC Research Institute
4 Independence Way
Princeton, NJ 08540
grove@research.nj.nec.com

**Joseph Y. Halpern*** 
Cornell University
Dept. of Computer Science
Ithaca, NY 14853
halpern@cs.cornell.edu
http://www.cs.cornell.edu/home/halpern

## Abstract

There are several well-known justifications for *conditioning* as the appropriate method for updating a single probability measure, given an observation. However, there is a significant body of work arguing for *sets* of probability measures, rather than single measures, as a more realistic model of uncertainty. Conditioning still makes sense in this context—we can simply condition each measure in the set individually, then combine the results—and, indeed, it seems to be the preferred updating procedure in the literature. But how justified is conditioning in this richer setting? Here we show, by considering an axiomatic account of conditioning given by van Fraassen, that the single-measure and sets-of-measures cases are very different. We show that van Fraassen's axiomatization for the former case is nowhere near sufficient for updating sets of measures. We give a considerably longer (and not as compelling) list of axioms that together force conditioning in this setting, and describe other update methods that are allowed once any of these axioms is dropped.

## 1 INTRODUCTION

A common criticism of the use of probability theory is that it requires the agent to make unrealistically precise uncertainty distinctions. One widely-used approach to dealing with this has been to consider sets of probability measures as a way of modeling uncertainty (see, for example, [Breese and Fertig 1991; Gilboa and Schmeidler 1993; Huber 1980; Kyburg 1974; Levi 1980; Smith 1961]). Given that one adopts the sets-of-measures model, how should one update these measures in the light of new evidence? There is an "obvious" approach available, which is to apply standard probabilistic conditioning to each of the measures in the set individually and then combine the results. It is typically taken for granted that this is the appropriate thing to do (see, for example, [Cozman 1997]). But what justifies this approach?

There have been numerous attempts to justify conditioning as the appropriate way to update single probability measures. The standard approach involves Dutch Book arguments [Kemeny 1955; Shimony 1955; Teller 1973]. However, these arguments have not always been viewed as so convincing; see Bacchus, Kyburg, and Thaler [1990] and Howson and Urbach [1989] for a summary of these arguments and some counter-arguments against them. In any case, even if we accept the standard justifications for conditioning, there is no *a priori* reason to believe that they must also apply to the sets-of-measures case. In fact, they may not, and demonstrating this is a major point of this paper. We focus here on a different, yet simple and compelling defense of (ordinary) conditioning, due to van Fraassen [van Fraassen 1987; Hughes and van Fraassen 1985]. Van Fraassen considers two simple and arguably quite reasonable properties that we might demand of an update process and shows that conditioning is the only mechanism that satisfies these properties. We show that these properties are not sufficient in the sets-of-measures case. Indeed, there are numerous other update mechanisms that satisfy them. We also show that, by postulating enough extra properties, we can recover conditioning as the unique solution; however, the properties we seem to need are far less compelling than those required for the original result.

We begin with an informal description of van Fraassen's result. He wants to examine arbitrary approaches for updating probabilities in the light of new evidence. Thus, he considers a function *upd* (for *update*) that takes two arguments—a probability measure Pr on a domain $W$ and a subset $B \subseteq W$—and returns a new probability measure *upd*(Pr, $B$), which, intuitively, is the result of updating Pr by the evidence $B$. It certainly seems reasonable to require

$$\text{if } \Pr' = \textit{upd}(\Pr, B), \text{ then } \Pr'(B) = 1. \qquad (1)$$

*This work was supported in part by NSF under grant IRI-96-25901 and by the Air Force Office of Scientific Research under grant F49620-96-1-0323.

That is, after updating, we should ascribe probability 1 to the evidence we have obtained.

Another reasonable principle to require is what van Fraassen calls *symmetry*. We can also think of it as *representation independence*, in the sense of [Halpern and Koller 1995]. Intuitively, suppose we represent a situation using the worlds in $W'$ rather than those in $W$. Let $f$ transform $W$ to $W'$; there is a corresponding transformation $f^*$ of a probability Pr on $W$ to a probability $f^*(\text{Pr})$ on $W'$. Then we would expect *upd* to respect this transformation. Roughly speaking, this means that if $\text{Pr}' = upd(\text{Pr}, B)$, then we would expect $f^*(\text{Pr}') = upd(f^*(\text{Pr}), f(B))$. More precisely, if $B \subseteq W'$ and $\text{Pr}' = upd(\text{Pr}, f^{-1}(B))$, then we would expect $f^*(\text{Pr}') = upd(f^*(\text{Pr}), B)$. The formal definition is given in Section 3, but an example here might help explain the intuition.

**Example 1.1:** Consider two agents who are reasoning about a given situation. One uses the primitive propositions $p$, $q$, and $r$; the second uses $p$ and $q$. Let $W$ consist of the eight truth assignments to $p$, $q$, and $r$, and let $W'$ consist of the four truth assignments to $p$ and $q$. We take the eight worlds in $W$ to be of the form $w_{ijk}$, $i, j, k \in \{0, 1\}$, where $i$, $j$, and $k$ give the truth value of $p$, $q$, and $r$, respectively. Thus, for example, in $w_{101}$, $p$ and $r$ are true while $q$ is false. Similarly, we take the worlds in $W'$ to have the form $w'_{ij}$. We now consider the obvious mapping $f$ from $W$ to $W'$ that maps $w_{ijk}$ to $w'_{ij}$. Given a measure Pr on $W$, $f$ induces a measure $f^*(\text{Pr})$ on $W'$ in the natural way, i.e., by taking $f^*(\text{Pr})(w'_{ij}) = \text{Pr}(\{w_{ij0}, w_{ij1}\})$. We want it to be the case that our updating rule respects this transformation. In this case, that would mean that $f^*(upd(\text{Pr}, f^{-1}(B))) = upd(f^*(\text{Pr}), B)$ for each $B \subseteq W'$. Thus, for example, we would have $f^*(upd(\text{Pr}, \{w_{ij0}, w_{ij1}\})) = upd(f^*(\text{Pr}), \{w_{ij}\})$. ∎

Van Fraassen showed that the only updating rule that is representation independent in this sense and satisfies (1) is conditioning.

We can apply van Fraassen's approach to the sets-of-measures case in a straightforward way. We again consider update functions that take two arguments, but now the first argument is a set of probability measure rather than a single probability measure, and the output is also a set of probability measures. Van Fraassen's two postulates have obvious analogues in this setting (which we formalize in Section 3). However, as we show by example, they are no longer strong enough to characterize conditioning.

One interesting update function that satisfies both conditions is what Voorbraak [1996] has called *constraining*. This updating function is defined as follows: given a set $X$ of probability measures and an observation $B$, it returns all the measures in $X$ that assign $B$ probability 1. That is, the observation $B$ is viewed as placing a constraint on the set of probability measure—namely, that $B$ must be assigned probability 1. We then return all the measures in the set that satisfy this new constraint. Voorbraak argues that constraining is actually more appropriate than conditioning when it comes to capturing a probabilistic analogue of the notion of *expansion* in the AGM [Alchourrón, Gärdenfors, and Makinson 1985] theory of belief change (where expansion is how beliefs change when we get extra information that is consistent with previously-held beliefs).

In Section 3, we provide seven postulates on update functions that suffice to guarantee that an update function on sets of measures acts like conditioning. Besides van Fraassen's postulates, our postulates include a "homomorphism" postulate, which says that the result of updating a set $X$ of measures is the union of the result of updating each element in $X$ separately, and a compositionality postulate, which says that updating by $B$ and then by $C$ is the same as updating by $B \cap C$ (and hence also the same as updating by $C$ and then by $B$). We also include a postulate that limits the amount by which the post-update probability of an event can exceed the value it would obtain under conditioning. The intuition for this is that an extremely improbable event should not receive a post-update probability that is "too large". The postulate that we use to capture this intuition is arguably too strong; it is an open problem to what extent it can be weakened.

Although our postulates are quite strong, we show that no subset of them suffices to force conditioning. Interestingly, if we drop our last postulate, then there are exactly two update functions that are consistent with the remaining six: conditioning and constraining.

We believe we are the first to try to axiomatize the updating of sets of probability measures, but others have certainly examined the issue of updating other notions of uncertainty that are related to sets of probability measures. Besides the work of Voorbraak cited above, we briefly mention three other lines of research:

- It is well known that a Dempster-Shafer belief function Bel [Shafer 1976] can be associated with the set of probability measures that dominate it, that is, the set $\mathcal{P}_{\text{Bel}} = \{\text{Pr} : \text{Pr}(A) \geq \text{Bel}(A) \text{ for all } A\}$. In fact, $\text{Bel}(A) = \inf_{\text{Pr} \in \mathcal{P}_{\text{Bel}}} \text{Pr}(A)$. One way of defining the update of Bel by a set $B$, considered in [Fagin and Halpern 1991; Jaffray 1992], is to take $\text{Bel}(\cdot|B) = \inf_{\text{Pr} \in \mathcal{P}_{\text{Bel}}} \text{Pr}(A|B)$. This approach to updating is quite different from Dempster's Rule of Conditioning [Shafer 1976]. (See [Halpern and Fagin 1992] for a discussion of the differences.) Moral and de Campos [1991] consider yet other approaches to updating belief functions.

- Gilboa and Schmeidler [1993] consider update rules for *non-additive probabilities* (of which belief functions are a special case, as are convex sets of probability measures). They show that under certain assumptions, the *maximum-likelihood* up-date rule is equivalent to Dempster's Rule of Conditioning. We discuss these results in more detail in Section 3.

- Walley [1991] has a theory of lower and upper *previsions* based on *gambles* and considers an approach to updating previsions called the *generalized Bayes rule*, which, as the name suggests, generalizes standard conditional probability. Sets of gambles can be associated with (convex) sets of probability measures; Moral and Wilson [1995] consider approaches to revising closed sets of gambles given another gamble[1] and relate their approaches to the AGM postulates.

The rest of this paper is organized as follows. In Section 2 we define update functions carefully and give some examples of them. In Section 3, we state our postulates. In Section 4 we outline the proof of our main result, which is that our postulates characterize conditioning. Despite the strength of our postulates, our proof is surprisingly difficult, which is perhaps further evidence that quite strong postulates are necessary to characterize conditioning in the sets-of-measures case. We conclude in Section 5.

## 2 UPDATE FUNCTIONS

The general framework we work in is a straightforward extension of van Fraassen's. Suppose we have a measure space $\mathcal{M} = (W, \mathcal{F})$, that is, a domain $W$ and an algebra $\mathcal{F}$ over $W$.[2] Let $\Delta_\mathcal{M}$ consist of all probability measures over $\mathcal{M}$. An *update function on* $\mathcal{M}$ is a function $Upd : 2^{\Delta_\mathcal{M}} \times \mathcal{F} \to 2^{\Delta_\mathcal{M}}$, such that $Upd(X, B) = \emptyset$ if $\Pr(B) = 0$ for all $\Pr \in X$.[3] That is, $Upd$ takes as input a set of probability measures over $\mathcal{M}$ and an element of $\mathcal{F}$, and returns a set of measures over $\mathcal{M}$. Intuitively, $Upd(X, B)$ consists of the result of updating the measures in $X$ by the observation $B$. If $X$ is the singleton set $\{\Pr\}$, we write $Upd(\Pr, B)$ rather than $Upd(\{\Pr\}, B)$. Note that for us, however, unlike for van Fraassen, $Upd(\Pr, B)$ is a set of measures (possibly empty), not a single measure.

Van Fraassen's symmetry requirement (i.e., representation independence), considers not one update function, but two, acting on different domains, and relates their outputs. Thus, we are interested in families $Upd^\mathcal{M}$ of update functions, one for each measure space $\mathcal{M}$. We use $Upd$ as a way of denoting the whole family $\{Upd^\mathcal{M}\}$.[4]

As defined, families of update functions can be completely arbitrary. They can act like conditioning in one space and return a fixed probability measure in another. We now give examples of seven families of update functions that are not completely arbitrary, in that they satisfy a number of properties of interest to us, although in some cases their behavior is quite far from conditioning.

- $Upd^\mathcal{M}_{cond}(X, B) = \{\Pr(\cdot|B) : \Pr \in X, \Pr(B) > 0\}$.
  $Upd_{cond}$ is the standard update via conditioning. More precisely, we condition when possible; we simply discard those probability measures $\Pr \in X$ such that $\Pr(B) = 0$.

- $Upd^\mathcal{M}_{constrain}(X, B) = \{\Pr \in X : \Pr(B) = 1\}$.
  $Upd_{constrain}$ is just Voorbraak's [1996] notion of constraining, as discussed in the Introduction. Note that $Upd^\mathcal{M}_{constrain}(X, B) = \emptyset$ if $X$ contains no probability measures $\Pr$ such that $\Pr(B) = 1$.

- $Upd^\mathcal{M}_{forget}(X, B) = \{\Pr \in \Delta_\mathcal{M} : \Pr(B) = 1\}$.
  With $Upd_{forget}$, we ignore the information in $X$ altogether. While this may seem to be a completely uninteresting update function, note that it can be viewed as modeling an agent who learns $B$, but then forgets what he knew before (which we can think of as being encoded by $X$). It points out the role of "no forgetting" in conditioning, an issue to which we return below.

- $Upd^\mathcal{M}_{trivial}(X, B) = \emptyset$.
  We have already seen that, in general, we may have $Upd(X, B) = \emptyset$ even if $X \neq \emptyset$. With $Upd_{trivial}$, we take this one step further and have the output be the empty set independent of $X$ and $B$. While this is clearly a rather uninteresting update function, we must be careful about what requirements to impose to ban it, so we do not ban too much.

- $Upd^\mathcal{M}_{closure}(X, B) = Upd^\mathcal{M}_{cond}(X^c, B) = (\{\Pr(\cdot|B) : \Pr \in X^c, \Pr(B) \neq 0\})$, where $X^c$ denotes the topological closure of $X$, that is, $X^c$ consists of all measures $\Pr$ such that for all $\epsilon > 0$, there exists a measure $\Pr' \in X$ such that $\sup_{A \in \mathcal{F}} |\Pr(A) - \Pr'(A)| < \epsilon$.
  $Upd_{closure}$ shows that update functions can take topological conditions into account. Note that $Upd_{cond}$ and $Upd_{closure}$ agree on all finite sets $X$.

---

[1] Since events can be viewed as a special case of gambles, this is a more general notion of updating than that considered here.

[2] An algebra $\mathcal{F}$ over $W$ is a set of subsets of $W$ that includes $W$ and is closed under complementation and union, so that if $A, B \in X$, then so are $\overline{A}$ and $A \cup B$. In the case that $W$ is infinite, we could also require that $\mathcal{F}$ be a $\sigma$-algebra, that is, closed under countable union. None of our results would change if we made this requirement.

[3] The final condition in this definition is analogous to the conventional restriction that one cannot condition on a measure 0 event. It is well known that the problem of defining a sensible notion of "update" for measure 0 events is a nontrivial one, even in the the conventional (single measure) framework. However, this problem is (largely) orthogonal to the topic of this paper. Note that this condition implies that $Upd(\emptyset, B) = \emptyset$.

[4] Readers concerned about cardinality considerations should think in terms of restricting to domains that have at most a certain cardinality, such as the cardinality of the reals.

The difference between them only arises if their first argument is infinite. For example, suppose $\mathcal{M}_2 = (\{1,2\}, 2^{\{1,2\}})$ and $X = \{\Pr \in \Delta_{\mathcal{M}_2} : \Pr(\{2\}) < 1\}$. Then $Upd_{cond}^{\mathcal{M}_2}(X, \{1,2\}) = X$, while $Upd_{closure}^{\mathcal{M}_2}(X, \{1,2\}) = \Delta_{\mathcal{M}_2}$, since $X^c = \Delta_{\mathcal{M}_2}$: every probability measure in $\Delta_{\mathcal{M}_2}$ not in $X$ must give $\{2\}$ probability 1, and can be approximated arbitrarily closely by a measure in $X$.

- $Upd_{subset}^W(\Pr, B) =$
$$\begin{cases} \{\Pr(\cdot | C \wedge B) : C \subseteq W, \ \Pr(C \wedge B) > 0\} & \text{if } \Pr(B) < 1 \\ \{\Pr\} & \text{if } \Pr(B) = 1 \end{cases}$$
$Upd_{subset}^W(X, B) = \cup_{\Pr \in X} Upd_{subset}^W(\Pr, B)$.

Intuitively, if $\Pr(B) < 1$, then $Upd_{subset}^W(\Pr, B)$ amounts to conditioning on all events we could learn in addition to $B$. We treat the case that $\Pr(B) = 1$ specially, to ensure that $Upd_{subset}$ satisfies one of the postulates we consider. For arbitrary sets $X$ of probability measures, we apply $Upd_{subset}$ pointwise (and then take unions). Our interest in $Upd_{subset}$ is motivated by the fact that it satisfies many natural properties while being quite different from $Upd_{cond}$ and $Upd_{constrain}$.

- $Upd_{ML}^W(X, B) = \{\Pr(\cdot | B) : \Pr \in X, \Pr(B) > 0, \Pr(B) = \sup_{\Pr' \in X} \Pr'(B)\}$.

$Upd_{ML}$ is the maximum likelihood rule considered by Gilboa and Schmeidler [1993] (except that they restrict to the case that $X$ is a closed convex set, which guarantees that there is some $\Pr \in X$ such that $\Pr(B) = \sup_{\Pr' \in X} \Pr'(B)$). It is an instance of what they call a *classical update rule*, which is one of the form $Upd^{\mathcal{M}}(X, B) = \{\Pr(\cdot | B) : \Pr \in X' \subseteq X\}$, for some appropriately chosen $X'$. Note that $Upd_{cond}$, $Upd_{constrain}$, $Upd_{trivial}$, and $Upd_{ML}$ are all classical update rules in this sense.

## 3 THE POSTULATES

What properties should an update function have? We want to start by imposing the two properties considered by van Fraassen. The first is easy to formalize in our framework.

P1. $Upd^{\mathcal{M}}(X, B) \subseteq \{\Pr \in \Delta_{\mathcal{M}} : \Pr(B) = 1\}$

That is, if we learn $B$, we want to assign probability 1 to $B$. Notice that all seven update functions described above satisfy this postulate.

To define the second postulate (i.e., representation independence) carefully, we review some material from [Halpern and Koller 1995]. What does it mean to shift from a representation (i.e., a measure space) $\mathcal{M} = (W, \mathcal{F})$ to another representation $\mathcal{M}' = (W', \mathcal{F}')$? There are many ways of shifting from one representation to another. For us, it suffices to consider what is perhaps the simplest case, where each world in $W'$ is associated with several worlds in $W$. We can think of representation $W$ as being richer than representation $W'$, in the sense of using additional primitive propositions or random variables to describe a world. This is the situation in Example 1.1, where in $W$ we used three primitive propositions to describe a world, whereas in $W'$ we used only two. We can then associate with each world in $W'$ all the worlds in $W$ that agree with it on all the primitive propositions it uses. Formally, this association is captured by a surjective map from $W$ to $W'$.

**Definition 3.1:** A *representation shift from* $\mathcal{M} = (W, \mathcal{F})$ *to* $\mathcal{M}' = (W, \mathcal{F}')$, also called an $\mathcal{M}$–$\mathcal{M}'$ *representation shift*, is a measurable surjective map from $W$ to $W'$, that is, a surjection $f : W \to W'$ such that $f^{-1}(B) \in \mathcal{F}$ for all $B \in \mathcal{F}'$ (where, as usual, $f^{-1}(B) = \{x \in W : f(x) \in B\}$). ∎

As is well known, $f^{-1}$ is a homomorphism with respect to unions and complementation, that is, $f^{-1}(B \cup B') = f^{-1}(B) \cup f^{-1}(B')$ and $f^{-1}(\overline{B}) = \overline{f^{-1}(B)}$ (where we use $\overline{U}$ to denote the complement of $U$). The fact that $f$ is surjective also makes $f^{-1}$ 1-1.[5] An $\mathcal{M}$–$\mathcal{M}'$ representation shift also induces a map $f^* : \Delta_{\mathcal{M}} \to \Delta_{\mathcal{M}'}$; we define $(f^*(\Pr))(A) = \Pr(f^{-1}(A))$.[6] Finally, if $X \subseteq \Delta_{\mathcal{M}}$, then we define $f^*(X) = \{f^*(\Pr) : \Pr \in X\}$.

With these definitions, we can formally state van Fraassen's representation independence property. Intuitively this says that, so long as we are updating by an event in $\mathcal{F}'$, it should not make any difference if we are fact working in a space $\mathcal{M}$ that is capable of making finer distinctions than does $\mathcal{M}'$. Another consequence is that the "labels" attached to points cannot affect how we update measures.

P2. Let $f$ be an $\mathcal{M}$–$\mathcal{M}'$ representation shift. If $X \subseteq \Delta_{\mathcal{M}}$ and $B \in \mathcal{F}'$, then $Upd^{\mathcal{M}'}(f^*(X), B) = f^*(Upd^{\mathcal{M}}(X, f^{-1}(B)))$.

As we said in the introduction (and will also follow from the results in Section 4), van Fraassen showed that if we consider update functions $upd$ from probability measures to probability measures (rather than from sets of probability measures to sets of probability measures), then P1 and P2 (appropriately modified to deal with $upd$ rather than $Upd$) suffice to guarantee that $upd$ is conditioning. However, it is easy to see that all seven of the update functions described in Section 2 satisfy both P1 and P2.

Can we impose other reasonable properties that restrict the set of allowable update functions? One property of conditioning is that order does not matter. Updating by $B$ and then $C$ is the same as updating by $C$ and then $B$, and both are the same as updating by $B \cap C$. This property does not follow from P1 and P2 in our more general setting. $Upd_{forget}$ provides a counterexample: if we update by $B$ and then by $C$ using $Upd_{forget}$, we get all the probability measures that give $C$ probability 1; if we update by $C$ and then $B$, then we get all the probability measures that give $B$ probability 1; if we update by $B \cap C$, we get all probability measures that give $B \cap C$ probability 1. Thus, we add the requirement that updates commute to our list of properties as well.

---

[5]In the language of [Halpern and Koller 1995], if $f$ is an $\mathcal{M}$–$\mathcal{M}'$ representation shift, then $f^{-1}$ is a faithful $\mathcal{M}'$–$\mathcal{M}$ embedding.

[6]$f^*$ is what van Fraassen calls a *measure embedding*.

P3. $Upd^{\mathcal{M}}(Upd^{\mathcal{M}}(X, B), C) = Upd^{\mathcal{M}}(X, B \cap C)$

Although P3 is a standard property of conditioning, it is far from innocuous. It can be viewed as encoding an assumption of "no forgetting". Intuitively, in order for updating by $B$ and then $C$ to be the same as updating by $B \cap C$, the agent must remember the information in $B$ when he is updating by $C$.

It is easy to see that P3 is not satisfied by either $Upd_{forget}$ or $Upd_{ML}$, although it is satisfied by the other update rules defined in Section 2. As observed in [Fagin and Halpern 1991; Jaffray 1992], the update rule for belief functions define by $\mathrm{Bel}(\cdot|B) = \inf_{\mathrm{Pr} \in \mathcal{P}_{\mathrm{Bel}}} \mathrm{Pr}(A|B)$ also does not satisfy P3.[7] On the other hand, Dempster's Rule of Conditioning does satisfy P3. Gilboa and Schmeidler [1993] provide sufficient conditions on $X$ that guarantee that $Upd_{ML}$ satisfies P3.[8] Moreover, they show $Upd_{ML}(X, B)$ acts like Dempster's Rule of Conditioning for those sets $X$ that satisfy these conditions.

We clearly need further postulates to rule out functions besides $Upd_{ML}$ and $Upd_{forget}$. The next postulate says our beliefs don't change if we learn information that we expected to be true all along (i.e., which was given probability 1 by all measures in our current set). This suffices to rule out $Upd_{trivial}$.

P4. $Upd^{\mathcal{M}}(X, B) = X$ if $\Pr(B) = 1$ for all $\Pr \in X$.

It is easy to see that P4 is satisfied by $Upd_{subset}$ as well as $Upd_{cond}$ and $Upd_{constrain}$. (Note that the special treatment of $Upd_{subset}(\Pr, B)$ in the case that $\Pr(B) = 1$ was necessary to ensure this.) Although P4 is not satisfied by $Upd_{closure}$, we could modify $Upd_{closure}(X, B)$ in the special case that $\Pr(B) = 1$ for all $\Pr \in X$ so that it does satisfy P4. We thus need a stronger condition to rule out update functions such as $Upd_{closure}$. The next postulate, which says that the action of an update function on a set of measures is determined by its action on the individual members of the set, does that.

---

[7]Jaffray [1992, Corollary 2] characterizes the restricted circumstances when P3 holds for this update rule.

[8]Suppose that $X$ consist of a set of probability measures on $\mathcal{M} = (W, \mathcal{F})$. Let $f_X(A) = \min\{\Pr(A) : \Pr \in X\}$ for $A \in \mathcal{F}$. The Gilboa-Schmeidler conditions say that (1) $X$ is convex, (2) $X = \{\Pr \in \Delta_{\mathcal{M}} : \Pr(A) \geq f_X(A)\}$, and (3) $f_X(A \cup B) + f_X(A \cap B) \geq f_X(A) + f_X(B)$, for all $A, B \in \mathcal{F}$. See [Gilboa and Schmeidler 1993] for the motivation for these conditions.

P5. $Upd^{\mathcal{M}}(X, B) = \cup_{\Pr \in X} Upd^{\mathcal{M}}(\Pr, B)$.

As we have seen, $Upd_{closure}$ does not satisfy P5; it acts like $Upd_{cond}$ on finite sets, but disagrees with $Upd_{cond}$ in general on arbitrary sets. $Upd_{ML}$ does not satisfy P5 either. It might seem that once we force the behavior of an update function to depend only on its behavior of singletons, we should be able to appeal immediately to van Fraassen's result. This, however, is not true, because $Upd^{\mathcal{M}}(\Pr, B)$ can still be an arbitrary set of probability measures. All of the update functions in our examples other than $Upd_{closure}$ and $Upd_{ML}$ satisfy P1, P2, and P5.

It might seem that P5 gives too special a role to the action of Upd on singleton sets. This is not in keeping with the spirit modeling uncertainty by arbitrary sets of probability measures. We can rewrite P5 to avoid mention of singleton sets by requiring instead that Upd commute with arbitrary unions, that is,

$$Upd^{\mathcal{M}}(\cup_{j \in J} X_j, B) = \cup_{j \in J} Upd^{\mathcal{M}}(X_j, B),$$

where $J$ is an arbitrary index set. It is easy to see that this postulate is equivalent to P5 while not giving a special role to singleton sets. We wrote P5 as we did because in fact we need it only for singleton sets. Note that it is important to allow the index set $J$ to be arbitrary here. The perhaps more appealing postulate $Upd^{\mathcal{M}}(X \cup Y, B) = Upd^{\mathcal{M}}(X, B) \cup Upd^{\mathcal{M}}(Y, B)$ (which can be extended by induction to show that Upd commutes with finite unions) is not strong enough to eliminate $Upd_{closure}$.

All of the properties we have considered so far are satisfied by $Upd_{subset}$. We consider here two ways of eliminating $Upd_{subset}$. Neither is as clean as we would like. After introducing the postulates, we discuss possible alternatives.

One way of eliminating $Upd_{subset}$ is to require that on a singleton argument, Upd returns either a singleton or the empty set.[9] More precisely, we have

P6'. $|Upd^{\mathcal{M}}(\Pr, B)| \leq 1$.

P6' again puts more emphasis on singleton sets than we would like, and seems somewhat strong. A quite different approach to eliminating $Upd_{subset}$ is based on the observation that it is not "continuous": every event in $B$ that has nonzero probability according to Pr (including ones whose probability is negligible) will be given full belief (probability 1) in at least one of the post-update measures, while the rest of $B$ (perhaps containing almost all of $B$'s probability according to Pr) is given probability 0. Perhaps there should be some limit on how much the probability of a small event can increase. The next postulate ensures this, by requiring an upper bound on the post-update probability relative to the original conditional probability.

---

[9]We must allow it to return an empty set since $Upd^{\mathcal{M}}(\Pr, B)$ is required to be $\emptyset$ if $\Pr(B) = 0$.

P6″. For all $\mathcal{M} = (W, \mathcal{F})$ and $\Pr \in \Delta_\mathcal{M}$, there exists a constant $c$ such that for all $A, B \in \mathcal{F}$ and $\Pr' \in Upd^\mathcal{M}(\Pr, B)$, we have $\Pr'(A) \leq c\Pr(A|B)$.

Clearly P6″ suffices to eliminate $Upd_{subset}$. However, it does not seem as natural as our other assumptions. Even if we accept the need for a continuity postulate, there seem to be weaker and more natural formalizations of it. In fact, the following postulate seems to assert continuity more directly:

P6*. For all $\mathcal{M} = (W, \mathcal{F})$, $\Pr \in \Delta_\mathcal{M}$, $B \in \mathcal{F}$, and $\epsilon > 0$, there exists $\delta > 0$ such that if $A \in \mathcal{F}$ and $\Pr(A|B) < \delta$, then $\Pr'(A) < \epsilon$ for every $\Pr' \in Upd^\mathcal{M}(\Pr, B)$.

P6* also suffices to rule out $Upd_{subset}$. However, we have not been able to prove that P1–P5 and P6* force $Upd_{cond}$ and $Upd_{constrain}$. Replacing P6* by P6′ or P6″ does the trick though. Let P6 state that either P6′ or P6″ holds.

P6. Either P6′ holds for all measures spaces $\mathcal{M}$ or P6″ holds for all measure spaces $\mathcal{M}$.

The main result of this paper, proved in the next section, is that $Upd_{cond}$ and $Upd_{constrain}$ are the only updating functions that satisfy P1–P6.

What is the relationship between P6′, P6″, and P6*? It is easy to see that P6″ implies P6*: given $B$, $\epsilon$, and $c$ as in P6″, we can take $\delta < \epsilon\Pr(B)/c$. As we show in the appendix, P6′ together with P1 and P2 implies both P6* and P6″. Finally, it follows from the main result of this paper that P6″ together with P1–P5 implies P6′ and P6*.

Once we are down to $Upd_{cond}$ and $Upd_{constrain}$, it is easy to add another postulate to get just $Upd_{constrain}$. The following weak postulate suffices:

P7. There exists some measure space $\mathcal{M} = (W, \mathcal{F})$, some set $X \subseteq \Delta_\mathcal{M}$, and some set $B \in \mathcal{F}$ such that $\Pr(B) \neq 1$ for all $\Pr \in X$ and $Upd^\mathcal{M}(X, B) \neq \emptyset$.

It should be clear that $Upd_{cond}$ satisfies P7, while $Upd_{constrain}$ does not.

## 4    THE MAIN THEOREM

The main result of the paper is the following.

**Theorem 4.1:** *The only update functions that satisfy P1–P6 are $Upd_{cond}$ and $Upd_{constrain}$.*

The following corollary is then immediate:

**Corollary 4.2:** *The only update function that satisfies P1–P7 is $Upd_{cond}$.*

In this section, we give a high-level outline of the proof of Theorem 4.1. Further details of the proof are deferred to the appendix. We omit the proof of some of the more technical and difficult lemmas because of limited space.

It is worth beginning with the following lemma, of which van Fraassen's result is a corollary. Roughly speaking, it says that the post-update probability we give to an event cannot be consistently smaller than its conditional probability.

**Proposition 4.3:** *Suppose that Upd satisfies P1 and P2 and that for some $\Pr' \in Upd^\mathcal{M}(\Pr, B)$ and $A$ such that $0 < \Pr(A|B) < 1$, we have $\Pr'(A) < \Pr(A|B)$. Then there also exists some $\Pr'' \in Upd^\mathcal{M}(\Pr, B)$ such that $\Pr''(A) > \Pr(A|B)$.*

**Proof:** See the appendix. ∎

Van Fraassen's result follows almost immediately from Proposition 4.3, as the following result shows.

**Proposition 4.4:** *If Upd satisfies P1 and P2, and $Upd^\mathcal{M}(\Pr, B) = \{\Pr'\}$, then $\Pr' = \Pr(\cdot|B)$.*

**Proof:** Suppose $Upd^\mathcal{M}(\Pr, B) = \{\Pr'\}$ and $\Pr'(A) \neq \Pr(A|B)$ for $0 < \Pr(A|B) < 1$. Then either $\Pr'(A) < \Pr(A|B)$ or $\Pr'(B - A) < \Pr(B - A|B)$. But this contradicts Proposition 4.3, because there is no corresponding $\Pr''$. A separate, but simple, argument is needed when $\Pr(A|B) \in \{0, 1\}$. If $\Pr(A|B) = 1$, consider any disjoint $A_1, A_2$ such that $A = A_1 \cup A_2$ and $0 < \Pr(A_1|B) < 1$. (We can assume that such $A_1, A_2$ exist, appealing to P2 if necessary.) But then

$$\begin{aligned}\Pr'(A) &= \Pr'(A_1) + \Pr'(A_2) \\ &= \Pr(A_1|B) + \Pr(A_2|B) \\ &= \Pr(A|B),\end{aligned}$$

where we use the fact that $0 < \Pr(A_1|B), \Pr(A_2|B) < 1$ and the previous argument. Finally, the case of $\Pr(A|B) = 0$ follows by considering $B - A$. Thus $\Pr' = \Pr(\cdot|B)$. ∎

It follows from Proposition 4.4 that P6′ implies both P6″ and P6* in the presence of P1 and P2.

The next step towards our result is to characterize $Upd_{constrain}$.

**Proposition 4.5:** *Suppose that Upd satisfies P1–P5 and that $Upd^\mathcal{M}(\Pr, B) = \emptyset$ for some space $\mathcal{M}$ and some $B$ with $\Pr(B) \neq 0$. Then $Upd = Upd_{constrain}$.*

**Proof:** See the appendix. ∎

Note that by P4, we cannot have $\Pr(B) = 1$ for the set $B$ in Proposition 4.5. Thus, Proposition 4.5 tells us that if Upd satisfies P1–P6 (in fact, even if it satisfies just P1–P5) and does *not* satisfy P7, then it must be $Upd_{constrain}$. It remains to show that if Upd satisfies P7 as well as P1–P6, then it must be $Upd_{cond}$.

In the case that $Upd$ satisfies P1–P5, P6', and P7, it follows from Proposition 4.5 that we must have $|Upd^{\mathcal{M}}(Pr, B)| = 1$ if $\Pr(B) \neq 0$. The fact that $Upd = Upd_{cond}$ now follows immediately from Proposition 4.4. Thus, it remains to show that if if $Upd$ satisfies P1–P5, P6'', and P7, then $Upd = Upd_{cond}$.

To show this, we first prove an easy lemma, which shows that $Upd$ must agree with conditioning at least on events with extreme probabilities.

**Lemma 4.6:** *If $Upd$ satisfies P1, P3, and P4, $\Pr \in \Delta_{\mathcal{M}}$, $A \subseteq B$, and $\Pr(A|B) = 1$ (resp., $\Pr(A|B) = 0$), then for all $\Pr' \in Upd^{\mathcal{M}}(\Pr, B)$, we have $\Pr'(A) = 1$ (resp., $\Pr'(A) = 0$).*

**Proof:** Let $M = (W, \mathcal{F})$. It suffices to prove the result for $\Pr(A|B) = 1$, because the other case follows by considering $B - A$.

Let $C = W - (B - A)$. Since $\Pr(A|B) = 1$, we have $\Pr(C) = 1$. By P4, we have $\{\Pr\} = Upd^{\mathcal{M}}(\Pr, C)$, and thus $Upd^{\mathcal{M}}(\Pr, B) = Upd^{\mathcal{M}}(Upd^{\mathcal{M}}(\Pr, C), B)$. By P3, $Upd^{\mathcal{M}}(Upd^{\mathcal{M}}(\Pr, C), B) = Upd^{\mathcal{M}}(\Pr, B \cap C) = Upd^{\mathcal{M}}(\Pr, A)$. It follows that $Upd^{\mathcal{M}}(\Pr, B) = Upd^{\mathcal{M}}(\Pr, A)$. But by P1, if $\Pr' \in Upd^{\mathcal{M}}(\Pr, A)$, then $\Pr'(A) = 1$. The result follows. ∎

We now introduce a key concept. Given $Upd$, a measure space $\mathcal{M} = (W, \mathcal{F})$, a measure $\Pr \in \Delta_{\mathcal{M}}$, and events $A \subseteq B \in \mathcal{F}$ such that $0 < \Pr(A) \le \Pr(B)$ define

$$U^{Upd, \mathcal{M}, \Pr}(A, B) = \sup_{\Pr' \in Upd^{\mathcal{M}}(\Pr, B)} \frac{\Pr'(A)}{\Pr(A)/\Pr(B)}.$$

(We take $U^{Upd, \mathcal{M}, \Pr}(A, B) = -\infty$ if $Upd^{\mathcal{M}}(\Pr, B) = \emptyset$.) The intuition is that $U^{Upd, \mathcal{M}, \Pr}(A, B)$ should be viewed as a measure of how far $Upd$ is from $Upd_{cond}$. It is the sup, over all the measures in $Upd^{\mathcal{M}}(\Pr, B)$, of the ratio of the (post-update) probability of $A$ to $\Pr(A|B)$. Of course, if $Upd = Upd_{cond}$, then $U^{Upd, \mathcal{M}, \Pr}(A, B) = 1$ for all $A$, $B$. As the following result shows, the converse holds as well.

**Proposition 4.7 :** *If $Upd$ satisfies P1–P4 and $U^{Upd, \mathcal{M}, \Pr}(A, B) = 1$ for all $\mathcal{M}$, $A$, and $B$, then $Upd = Upd_{cond}$.*

**Proof:** Suppose that $Upd \neq Upd_{cond}$. Then there exists a measure space $\mathcal{M} = (W, \mathcal{F})$, $\Pr \in \Delta_{\mathcal{M}}$, $B \in \mathcal{F}$, $A \subseteq B$, and $\Pr' \in Upd^{\mathcal{M}}(\Pr, B)$ such that $\Pr'(A) \neq \Pr(A|B)$. By Lemma 4.6 we know that $\Pr(A|B)$ is neither 0 nor 1. If $\Pr'(A) > \Pr(A|B)$, then $U^{Upd, \mathcal{M}, \Pr}(A, B) > 1$, contradicting our assumption. But otherwise, we have $\Pr'(A) < \Pr(A|B)$ and so, by Proposition 4.3, there exists $\Pr'' \in Upd^{\mathcal{M}}(\Pr, B)$ such that $\Pr''(A) > \Pr(A|B)$. Again, this contradicts our assumption. ∎

In light of Proposition 4.7, we can complete the proof of Theorem 4.1 by proving the following result:

**Proposition 4.8:** *If $Upd$ satisfies P1–P5, P6'', and P7, then $U^{Upd, \mathcal{M}, \Pr}(A, B) = 1$ for all $\mathcal{M}$, $A$, $B$.*

Despite the strength of P6'', the proof of Proposition 4.8 turns out to be surprisingly difficult. (More accurately, we have not been able to find a proof that is not difficult!) The details are in the appendix.

## 5   DISCUSSION

The main purpose of this paper is to illustrate how different the set-of-measures model can be, technically, from the standard single-measure model. In general, it is unwise to simply assume that a result in the standard model can be trivially "lifted" to apply in general.

In terms of understanding update procedures, one obvious next step would be to examine some of the other justifications for conditioning in the single-measure model, to see to what extent they carry over to the new setting. There are also outstanding questions even in the axiomatic framework considered here. P6, in particular, is quite a strong assumption. Is it really necessary? We conjecture that our main result actually holds with P6 replaced by P6*, although we have not been able to prove this. (Recall that P6' implies both P6'' and P6* in the presence of P1 and P2.)

We are clearly not advocating P6 (or even P6*) as being anywhere near as compelling as, say, P1 or P2. (Of course, one can raise reasonable arguments against P1 and P2—and most of the other postulates—as well.) So what does this say about $Upd_{cond}$ and $Upd_{constrain}$? It seems quite plausible to us that other update procedures, such as $Upd_{ML}$, that will be appropriate in some circumstances. We believe that a more careful investigation into such alternative rules, and a continued effort to try and clearly determine what makes an update rule appropriate to a given domain, would be worthwhile.

## A   APPENDIX: PROOFS

For the proofs, it is useful to define the family of spaces $\mathcal{M}_n = (\{1, \ldots, n\}, 2^{\{1, \ldots, n\}})$.

**Proposition 4.3:** *Suppose that $Upd$ satisfies P1 and P2 and that for some $\Pr' \in Upd^{\mathcal{M}}(\Pr, B)$ and $A$ such that $0 < \Pr(A|B) < 1$, we have $\Pr'(A) < \Pr(A|B)$. Then there also exists some $\Pr'' \in Upd^{\mathcal{M}}(\Pr, B)$ such that $\Pr''(A) > \Pr(A|B)$.*

**Proof:** Consider $\mathcal{M} = (W, \mathcal{F})$ and assume that $B \neq W$. (The argument if $B = W$ is almost identical and left to the reader.) By P2, it suffices to prove the result in the case that $\mathcal{M} = \mathcal{M}_3$. (For any other $\mathcal{M}$, we can consider the surjection that maps $A$ to 1, $B - A$ to 2, and $W - B$ to 3 and then appeal to P2.)

There are now two cases to consider. First, assume that $\Pr(A|B) < 1$ is rational, say $m/n$. Now consider the space $\mathcal{M}_{n+1}$. Let $A = \{1,\ldots,m\}$, $B = \{1,\ldots,n\}$, and Pr give each of $1,\ldots,n$ equal probability. We can consider the surjection $g : W_{n+1} \to W_3$ that maps $1,\ldots,m$ to 1, $m+1,\ldots,n$ to 2, and $n+1$ to 3. Thus, by P2 again, the result is true for $\mathcal{M}_3$ if we can show that it is true for an arbitrary measure Pr on $\mathcal{M}_{n+1}$ and $\Pr' \in Upd^{\mathcal{M}_{n+1}}(\Pr, B)$. Note that the reason we have to introduce both $\mathcal{M}_3$ and $\mathcal{M}_{n+1}$ is that there is always a mapping from any other $\mathcal{M}$ for which the proposition is relevant into the former space, but not necessarily into the latter.

Clearly $\Pr'$ does not give equal probability to each of the points $1,\ldots,n$ (for if it did, we would have $\Pr'(A) = m/n$). In fact, the average probability of a point in $A$ (according to $\Pr'$) is $1/n - \epsilon/m$, where $\epsilon = \Pr(A|B) - \Pr'(A)$. There are two cases to consider: If $\Pr(A|B) \geq 1/2$, then $n - m \leq m$. In this case, let $C$ consist of the $n - m$ elements of $A$ with the lowest probability (according to $\Pr'$). We must have $\Pr'(C) \leq (n-m)/n - (n-m)\epsilon/m$. Let $h$ be a permutation on $\{1,\ldots,n+1\}$ that switches the points in $C$ with the points in $B - A$ such that $h(n+1) = n+1$. Note that $h^*(\Pr) = \Pr$ (since Pr gives equal probability to all the points in $B$). Let $\Pr'' = h^*(\Pr')$. Note that $\Pr''(B - A) \leq (n-m)/n - ((n-m)/n)\epsilon$, so $\Pr''(A) \geq m/n + ((n-m)/n)\epsilon$. If $\Pr(A|B) < 1/2$, then $n - m > m$, and a similar argument works: This time, let $C$ consist of the $m$ elements of $A$ with lowest probability. In this case we get that $\Pr''(A) \geq m/n + (m/n)\epsilon$. Since $h^*(\Pr') \in Upd^{\mathcal{M}}(\Pr, B)$ by P2, we are done. Note that, in either case, $\Pr''(A) \geq \Pr(A|B) + \min(\Pr(A|B), 1 - \Pr(A|B))\epsilon$.

Next suppose that $\Pr(A|B)$ is irrational. Choose $r$ rational such that $\Pr(A|B) > r > \Pr(A|B) - \min(r, 1-r)\epsilon/2$. By using P2 again, it suffices to prove the result for $\mathcal{M}_4$, with $A = \{1,2\}$, $B = \{1,2,3\}$, and $\Pr(\{1\}|B) = r$. Let $A' = \{1\}$. Since $\Pr'(A') \leq \Pr'(A) = \Pr(A|B) - \epsilon$, it follows that $\Pr'(A') < r - \epsilon/2$. Since $\Pr(A'|B)$ is rational by construction, by the previous argument, there exists $\Pr'' \in Upd^{\mathcal{M}}(\Pr, B)$ such that $\Pr''(A') \geq r + \min(r, 1-r)\epsilon/2 > \Pr(A|B)$. Since $A' \subseteq A$, we have $\Pr''(A) > \Pr(A|B)$, as desired. ∎

**Proposition 4.5:** *Suppose that Upd satisfies P1–P5 and that $Upd^{\mathcal{M}}(\Pr, B) = \emptyset$ for some space $\mathcal{M}$ and $\Pr(B) \neq 0$. Then $Upd = Upd_{constrain}$.*

We prove this using Lemmas A.1–A.3.

**Lemma A.1:** *If Upd satisfies P2, $Upd^{\mathcal{M}}(\Pr, B) = \emptyset$, and $\Pr(B) = \alpha < 1$, then for every space $\mathcal{M}'$, probability measure $\Pr' \in \Delta_{\mathcal{M}'}$, and $B' \in \mathcal{F}'$ such that $\Pr'(B') = \alpha$, we have $Upd^{\mathcal{M}'}(\Pr', B') = \emptyset$.*

**Proof:** First assume that $\mathcal{M}' = \mathcal{M}_2$ $B' = \{2\}$. Then the result is immediate using P2. (Consider the $\mathcal{M}$-$\mathcal{M}'$ representation shift that maps $B$ to $\{2\}$ and $\overline{B}$ to $\{1\}$.) Now if $\mathcal{M}'$ is arbitrary, we again get the result by applying P2 and using the fact that it holds for $\mathcal{M}_2$. (Again, consider the $\mathcal{M}'$-$\mathcal{M}_2$ representation shift that maps $B'$ to $\{2\}$ and $\overline{B}$ to $\{1\}$.) ∎

We can bootstrap our way up to an even stronger lemma.

**Lemma A.2:** *If Upd satisfies P2 and P3, and $Upd^{\mathcal{M}}(\Pr, B) = \emptyset$ for some $B$ with $\Pr(B) = \alpha < 1$, then for every domain $\mathcal{M}'$, probability measure $\Pr' \in \Delta_{\mathcal{M}'}$, and $B' \subseteq W'$ such that $\Pr'(B') < \alpha$, we have $Upd^{\mathcal{M}'}(\Pr', B') = \emptyset$.*

**Proof:** First consider the space $\mathcal{M}_3$ and let $\Pr''$ be a measure such that $\Pr''(\{2,3\}) = \alpha$ and $\Pr''(\{3\}) = \beta < \alpha$. By Lemma A.1, we have that $Upd^{\mathcal{M}_3}(\Pr'', \{2,3\}) = \emptyset$. By P3, we have that

$$Upd^{\mathcal{M}_3}(\Pr'', \{2\}) = Upd^{\mathcal{M}_3}((Upd^{\mathcal{M}_3}(\Pr'', \{2,3\}), \{2\}))$$
$$= Upd^{\mathcal{M}_3}(\emptyset, \{2\}) = \emptyset.$$

By Lemma A.1 again, it follows that for every space $\mathcal{M}' = (W', \mathcal{F}')$, probability measure $\Pr' \in \Delta_{\mathcal{M}'}$, and $B' \in \mathcal{F}'$ such that $\Pr'(B') = \beta$, we have $Upd^{\mathcal{M}'}(\Pr', B') = \emptyset$. Since $\beta$ was arbitrary, the desired result follows. ∎

Finally, we get the strongest possible result of this type.

**Lemma A.3:** *If Upd satisfies P1, P2, P3, and P5, $Upd^{\mathcal{M}}(\Pr, B) = \emptyset$, then $Upd^{\mathcal{M}'}(\Pr', B') = \emptyset$ for every domain $\mathcal{M}'$, probability measure $\Pr' \in \Delta_{\mathcal{M}'}$, and set $B'$ such that $\Pr'(B') < 1$.*

**Proof:** Let $\gamma^* = \sup_{\mathcal{M}',B}\{\Pr'(B') : Upd^{\mathcal{M}'}(\Pr', B) = \emptyset\}$. We want to show that $\gamma^* = 1$. Suppose by way of contradiction that $\gamma^* < 1$. Choose $\epsilon > 0$ such that $0 < \gamma^* - \epsilon$, $\gamma^* + \epsilon < 1$, and $(\gamma^* - \epsilon)/(\gamma^* + \epsilon) > \gamma^*$. Consider the space $\mathcal{M}_3$ again and let $\Pr'$ be such that $\Pr'(2) = \gamma^* - \epsilon$ and $\Pr'(3) = 2\epsilon$. By choice of $\gamma^*$ and Lemma A.2, we have that $Upd^{\mathcal{M}_3}(\Pr', \{2\}) = \emptyset$ and $Upd^{\mathcal{M}_3}(\Pr', \{2,3\}) \neq \emptyset$. By P3, we have

$$Upd^{\mathcal{M}_3}((Upd^{\mathcal{M}_3}(\Pr', \{2,3\}), \{2\}) \qquad (2)$$
$$= Upd^{\mathcal{M}_3}(\Pr', \{2\}) = \emptyset.$$

However, since $Upd^{\mathcal{M}_3}(\Pr', \{2,3\}) \neq \emptyset$, by Proposition 4.3, there exists some $\Pr'' \in Upd^{\mathcal{M}_3}(\Pr', \{2,3\})$ such that $\Pr''(2) \geq \Pr'(\{2\}|\{2,3\}) = (\gamma^* - \epsilon)/(\gamma^* + \epsilon) > \gamma^*$. By Lemma A.2 and the choice of $\gamma^*$, we have that $Upd^{\mathcal{M}_3}(\Pr'', \{2\}) \neq \emptyset$. By P5, $Upd^{\mathcal{M}_3}(\Pr'', \{2\}) \subseteq Upd^{\mathcal{M}_3}((Upd^{\mathcal{M}_3}(\Pr', \{2,3\}), \{2\})$. This contradicts (2). Thus, we must have $\gamma^* = 1$. ∎

It follows from Lemmas A.2 and A.3 that $Upd^{\mathcal{M}}(\Pr, B) = \emptyset$ if $\Pr(B) < 1$. Proposition 4.5 now follows immediately from P4 and P5. ∎

This completes the proof of Proposition 4.5. In order to prove Theorem 4.1, it remains only to prove Proposition 4.8. We first need a few preliminary lemmas. The first shows that the value of $U^{Upd,\mathcal{M},\Pr}(A, B)$ depends only on $Upd$ and the values $\Pr(A)$ and $\Pr(B)$, but otherwise does not depend on the details of $\mathcal{M}$ or the exact identity of $A$ and $B$.

**Lemma A.4:** *If $Upd$ satisfies P1–P5 and P7 there is a function $V^{Upd}(x, y)$ defined for $0 < x \le y \le 1$ such that $U^{Upd,\mathcal{M},\Pr}(A, B) = V^{Upd}(\Pr(A), \Pr(B))$ for all $A \subseteq B \in \mathcal{F}$ with $\Pr(A) > 0$.*

**Proof:** For $0 < x < y < 1$, let $\Pr_{x,y} \in \Delta^{\mathcal{M}_3}$ be defined so that $\Pr_{x,y}(1) = x$ and $\Pr_{x,y}(2) = y - x$ (so that $\Pr_{x,y}(\{1, 2\}) = y$). Finally, define

$V^{Upd}(x, y) =$
$\begin{cases} 1 & \text{if } y = 1 \text{ or } x = y \\ U^{Upd,\mathcal{M}_2,\Pr_{x,y}}(\{1\}, \{1, 2\}) & \text{if } 0 < x < y < 1. \end{cases}$

We must show that this definition has the required properties. Consider any $\mathcal{M}, A, B, \Pr$ such that $A \subseteq B$. If $\Pr(A) < \Pr(B) < 1$, then it is immediate from P2 that $U^{Upd,\mathcal{M},\Pr}(A, B) = V^{Upd}(\Pr(A), \Pr(B))$. If $\Pr(A) = \Pr(B)$, then Lemma 4.6 assures us that if $\Pr' \in Upd^{\mathcal{M}}(\Pr, B)$ then $\Pr'(A) = 1$ and hence $U^{Upd,\mathcal{M},\Pr}(A, B) = 1$. Thus, the result follows as long as $Upd^{\mathcal{M}}(\Pr, B)$ is nonempty. But this follows from Proposition 4.5 and P7. Finally, if $\Pr(B) = 1$, the result is immediate from P4.[10] ∎

**Lemma A.5:** *If $Upd$ satisfies P1–P5 and P7, then for any fixed $y$, $V^{Upd}(x, y)$ is a non-increasing function of $x$ such that $V^{Upd}(y, y) = 1$.*

**Proof:** See the full paper. ∎

We need one more, rather technical, lemma. Our overall goal is to show that $Upd^{\mathcal{M}}(\Pr, C) = \{\Pr(\cdot|C)\}$. But perhaps it is reasonable to first ask a weaker question: is it true that $\Pr(\cdot|C) \in Upd^{\mathcal{M}}(\Pr, C)$? While the following lemma does not quite show this, it proves something in a very similar spirit. In particular, the lemma implies that, for every event $B \subset C$, there is some measure $\Pr^B \in Upd^{\mathcal{M}}(\Pr, C)$ such that $\Pr^B(B) = \Pr(B|C)$. Note, however, that $\Pr^B$ may depend on $B$ and so this does not prove that $\Pr(\cdot|C) \in Upd^{\mathcal{M}}(\Pr, C)$. On the other hand, the lemma also does something more than just assert the existence of $\Pr^B$. Consider another event $A$ disjoint from $B$ and any

---

[10]We remark that the only place we use P3 in this proof is in the appeal to Lemma 4.6. With a little more effort, the result can be proved even without P3.

$\Pr' \in Upd^{\mathcal{M}}(\Pr, C)$. Of course, $\Pr'(A)$ does not necessarily equal $\Pr(A|C)$. But the lemma shows that there is another measure $\Pr'' \in Upd^{\mathcal{M}}(\Pr, C)$ that agrees with $\Pr'$ on $A$ and "looks like conditioning" outside $A$, at least with respect to $B$. More precisely, $\Pr''(B)$ is exactly the final probability of $C - A$ (i.e., $1 - \Pr'(A)$) times the *prior* probability $\Pr(B|C - A)$. Note that if $A = \emptyset$ this reduces to the earlier claim (since we can then take $\Pr^B = \Pr''$).

**Lemma A.6:** *Suppose that $Upd$ satisfies P1, P2, P6''. For all $A, B \subset C \subset W$ such that $A$ and $B$ are disjoint elements of $\mathcal{F}$, $\Pr \in \Delta_{\mathcal{M}}$, $\Pr(B) > 0$, and $\Pr' \in Upd^{\mathcal{M}}(\Pr, C)$, there exists $\Pr'' \in Upd^{\mathcal{M}}(\Pr, C)$ such that $\Pr''(A) = \Pr'(A)$ and $\Pr''(B) = (1 - \Pr'(A)) \Pr(B|C - A)$.*

**Proof:** See the full paper. ∎

This omitted proof is rather complex. Unlike the other results in this section, which involved only finite spaces, it makes crucial of an uncountable space. More precisely, our proof makes crucial use of P2 applied to representation shifts involving an uncountable space (the unit interval with Lebesgue measure). We conjecture that the result would actually be false if we restrict P2 to representation shifts involving only countable spaces. Of course, our main theorem might still be true even if we restrict P2 to countable spaces; it might have a different proof that does not rely on Lemma A.6.

Another point worth noting is that Lemma A.6 can be proved using the weaker postulate P6* rather than P6''. Unfortunately, this does not seem to be true for Proposition 4.8, which we are finally ready to prove.

**Proposition 4.8:** *If $Upd$ satisfies P1–P5, P6'', and P7, then $U^{Upd,\mathcal{M},\Pr}(A, B) = 1$ for all $\mathcal{M}, A, B$.*

**Proof:** By Lemma A.5, we see that $V^{Upd}(x, y)$ is bounded below by 1 (and in fact the bound is attained when $x = y$). On the other hand, by P6'', there is also a finite upper bound $c$. In fact, we can assume that $c$ is the least upper bound. Suppose by way of contradiction that $c > 1$.

The basic idea of the proof is to use P3 and show that there is a sequence of two iterated updates in which some event's probability grows by more than $c$ (relative to what we would expect from conditioning). We also show (using P3) that there is a single update which would give the same result. However, this contradicts the definition of $c$.

Consider $c' = (c + \sqrt{c})/2$; note that $c'^2 > c$ but $1 < c' < c$. In particular, by the definition of $c$ and since $c' < c$ we can find some $x, z$ such that $V^{Upd}(x, z) > c'$. Furthermore, by the monotonicity of $V^{Upd}$, we also have $V^{Upd}(x', z) > c'$ for $x' < x$. In the following, consider any $x' < x$ such that $cx'/z \le x$. Now consider the space $\mathcal{M}_4$. Let $A = \{1\}$, $B = \{1,2\}$, $C = \{1,2,3\}$, and let Pr be a measure such that $\Pr(A) = x'$ and $\Pr(C) = z$. This does not completely specify Pr; in particular, the only constraint so far on $\Pr(B)$ is that it lies between $x'$ and $z$. Note, however, that the set $\{r : \Pr'(A) = r \text{ for some } \Pr' \in Upd^{\mathcal{M}_4}(\Pr, C)\}$ is independent of the exact probability we give to $B$, because of postulate P2. In the following, choose some value $r$ from this set such that $r > c'x'/z$; the definition of $V^{Upd}$ guarantees that such an $r$ must exist, since $V^{Upd}(x', z) > c'$. Note also that $r < cx'/z$ by definition of $c$. Thus, $r < x$, by choice of $x'$.

We now complete the specification of Pr by defining $\Pr(2) = (z - r)(z - x')/(1 - r)$. It is easily verified that $\Pr(\{1,2\})$ is then between $x'$ and $z$. Moreover, $\Pr(\{2\} | \{2,3\})$ is $(z - r)/(1 - r)$ (since $\Pr(\{2,3\}) = \Pr(C - A) = z - x'$). Therefore, by Lemma A.6, there is some $\Pr' \in Upd^{\mathcal{M}_4}(\Pr, C)$ such that not only is $\Pr'(A) = r$ but furthermore
$\Pr'(B) = \Pr'(1) + \Pr'(2) = r + (1-r)(z-r)/(1-r) = z$.

However, now consider $Upd^{\mathcal{M}_4}(\Pr', B)$. We know that $\Pr'(A) = r < cx'/z < x$ and $\Pr'(B) = z$. Therefore (by the monotonicity of $V^{Upd}$) we have $V^{Upd}(r, z) > c'$. Thus, there must be some $\Pr'' \in Upd^{\mathcal{M}_4}(\Pr', B)$ such that $\Pr''(A) > c'r/z > c'^2 x'/z^2$.

By P3, we know that conditioning on $C$ then $B$ must give the same result as if we were to condition directly on $B$. Thus, $Upd^{\mathcal{M}_4}(\Pr, B)$ must also contain $\Pr''$. It follows that $V^{Upd}(x', \Pr(B)) \geq \Pr''(A) \Pr(B)/x' \geq c'^2 \Pr(B)/z^2$. Recall that $c'^2 > c$. We thus will have a contradiction with the definition of $c$ if we can show there exists $x'$ such that $\Pr(B)/z^2$ is close enough to 1 so that $c'^2 \Pr(B)/z^2$ exceeds $c$. But, in fact, we know that $\Pr(B) = x' + (z - r)(z - x')/(1 - r)$. It is clear that by choosing $x'$ (and hence also $r$) to be sufficiently small, we can make this as close to 1 as we wish. The result follows. ∎

## Acknowledgements

We thank Nick Littlestone, Dale Schuurmans, and Bas van Fraassen for useful discussions.